\documentclass[12pt]{article}
\usepackage{graphicx}
\usepackage{amsmath}
\usepackage{amssymb}
\usepackage{physics}
\usepackage{authblk}
\usepackage{amsthm}
\usepackage{booktabs}
\usepackage{algorithm}
\usepackage{algpseudocode}

\usepackage{geometry}
\usepackage{setspace}
\usepackage{titlesec}
\usepackage[numbers]{natbib}  
\usepackage[colorlinks=true, linkcolor=black, citecolor=blue]{hyperref}

\titleformat{\section}[block]{\normalfont\Large\bfseries}{\thesection}{1em}{}
\titleformat{\subsection}[block]{\normalfont\large\bfseries}{\thesubsection}{1em}{}
\titleformat{\subsubsection}[block]{\normalfont\normalsize\bfseries}{\thesubsubsection}{1em}{}

\title{\textbf{Why Flow Matching is Particle Swarm Optimization?}}

\author[1]{\small Kaichen Ouyang\thanks{Corresponding author: \texttt{oykc@mail.ustc.edu.cn}}}
\affil[1]{\small Department of Mathematics, University of Science and Technology of China, Hefei 230026, China}
\date{}

\begin{document}
\maketitle
\vspace{-1em}  

\begin{abstract}
This paper preliminarily investigates the duality between flow matching in generative models and particle swarm optimization (PSO) in evolutionary computation. Through theoretical analysis, we reveal the intrinsic connections between these two approaches in terms of their mathematical formulations and optimization mechanisms: the vector field learning in flow matching shares similar mathematical expressions with the velocity update rules in PSO; both methods follow the fundamental framework of progressive evolution from initial to target distributions; and both can be formulated as dynamical systems governed by ordinary differential equations. Our study demonstrates that flow matching can be viewed as a continuous generalization of PSO, while PSO provides a discrete implementation of swarm intelligence principles. This duality understanding establishes a theoretical foundation for developing novel hybrid algorithms and creates a unified framework for analyzing both methods. Although this paper only presents preliminary discussions, the revealed correspondences suggest several promising research directions, including improving swarm intelligence algorithms based on flow matching principles and enhancing generative models using swarm intelligence concepts.
\end{abstract}


\section{Introduction}
Complex adaptive systems in nature exhibit remarkable capabilities of self-organization and emergent intelligence, spanning both physical and biological domains\cite{holland1995hidden,kauffman1992origins,mitchell2009complexity}. From the molecular dynamics governing fluid flows to the collective behaviors of bird flocks and ant colonies, these systems demonstrate how simple local interactions can generate sophisticated global patterns\cite{cross1993pattern,reynolds1987flocks,bonabeau1999swarm}. This rich tapestry of natural phenomena has inspired two major classes of computational models in machine learning: generative models that emulate physical processes, and evolutionary computation that mimic biological adaptation\cite{jo2023promise,eiben2015evolutionary}.

Generative models have successfully captured complex dynamics through mathematical abstraction. Diffusion models simulate thermodynamic processes of particle diffusion and reversal\cite{ho2020denoising}, while flow matching methods directly learn the vector fields that govern continuous transformations between distributions\cite{lipman2022flow}. These approaches share a common foundation in modeling physical systems through differential equations, whether stochastic (as in diffusion processes) or deterministic (as in flow matching).

Conversely, evolutionary computation draws inspiration from biological complex adaptive systems. Swarm intelligence algorithms, such as particle swarm optimization (PSO)\cite{tang2021review}, replicate the emergent coordination observed in animal groups, where simple rules about local interactions and global information sharing lead to sophisticated optimization behaviors. Similarly, evolutionary algorithms implement Darwinian principles of selection, recombination, and mutation to solve complex problems through simulated evolution \cite{zelinka2015survey}.

Despite their shared roots in complex systems theory, these two paradigms—generative models and evolutionary computation—have developed largely independently. This paper bridges this divide by revealing fundamental connections between flow matching in generative modeling\cite{lipman2022flow} and particle swarm optimization in evolutionary computation\cite{kennedy1995particle}. We demonstrate that these seemingly distinct approaches are in fact mathematical duals, with flow matching representing a continuous generalization of PSO's discrete population dynamics.

Our work makes three primary contributions: First, we establish mathematical equivalences between the vector fields in flow matching and the velocity update rules in PSO, showing how both methods optimize population distributions through iterative updates (Section 3). Second, we demonstrate that the ordinary differential equation formulation of flow matching provides a rigorous theoretical framework for analyzing PSO dynamics, particularly in understanding convergence properties. Third, we identify practical synergies between the approaches, suggesting how flow matching's gradient information could enhance PSO's optimization capabilities while maintaining its population-based exploration advantages.

The significance of this unification extends beyond theoretical interest. Recent challenges in both fields—such as mode collapse in generative models and premature convergence in evolutionary algorithms—may find solutions through cross-pollination of ideas. For instance, the continuous-time formulation of flow matching could help analyze and improve PSO's convergence guarantees, while PSO's diversity maintenance mechanisms might address limitations in flow matching for multimodal distributions.

The remainder of this paper is organized as follows: Section 2 reviews related work in both generative models and evolutionary computation. Section 3 presents our core theoretical analysis connecting flow matching and PSO. Section 4 discusses implications and future research directions emerging from this connection. Our conclusions highlight how this unified perspective opens new possibilities for algorithm development in both fields.

\section{Related Work}
\subsection{Generative Models}
Recent advancements in generative models have revolutionized artificial intelligence, enabling high-quality data synthesis across diverse domains. Two prominent approaches have emerged: diffusion models and flow matching models. 

\paragraph{Diffusion Models} 
Diffusion models have established themselves as a powerful generative approach through iterative denoising processes. Key developments include Denoising Diffusion Probabilistic Models (DDPM)\cite{ho2020denoising}, which laid the foundation for modern diffusion approaches, and Score-based Generative Models (SGMs)\cite{song2019generative} that formulate generation through stochastic differential equations. Recent innovations like Latent Diffusion Models (LDM)\cite{rombach2022high} operate in compressed latent spaces for improved computational efficiency, with Stable Diffusion becoming the most prominent open-source implementation for image generation.

\paragraph{Flow Matching Models}
Flow matching models offer an alternative paradigm by learning deterministic transformation paths. This category includes Continuous Normalizing Flows (CNF)\cite{chen2018neural} that utilize neural ODEs for density estimation, and Rectified Flows which enhance efficiency through trajectory straightening\cite{liu2022flow}. The field has also seen Flow Matching techniques employing simulation-free continuous objectives, as well as Optimal Transport Flow (OT-Flow) models that combine optimal transport theory with flow-based approaches\cite{lipman2022flow}. These methods collectively provide efficient alternatives to traditional generative modeling techniques.

\subsection{Evolutionary Computation}
Evolutionary computation techniques draw inspiration from biological evolution to address complex optimization challenges. Currently, two main approaches dominate the field: swarm intelligence algorithms that simulate collective behaviors of natural systems, and evolutionary algorithms that implement mechanisms of biological evolution.

\paragraph{Swarm Intelligence Algorithms}
Swarm intelligence algorithms emulate collective behaviors observed in decentralized natural systems. Three foundational approaches include Particle Swarm Optimization (PSO)\cite{kennedy1995particle}, which pioneered the simulation of social behavior in optimization; Ant Colony Optimization (ACO)\cite{dorigo2007ant} that introduced pheromone-based pathfinding for combinatorial problems; and the Firefly Algorithm (FA)\cite{yang2009firefly} that established light-intensity based attraction mechanisms for multimodal optimization.

\paragraph{Evolutionary Algorithms}
Evolutionary Algorithms implement biological evolution mechanisms for optimization. Three principal variants comprise Genetic Algorithms (GA)\cite{holland1992genetic} featuring chromosome crossover and mutation operations; Differential Evolution (DE)\cite{storn1997differential} utilizing vector differences for efficient global optimization; and Covariance Matrix Adaptation Evolution Strategy (CMA-ES)\cite{hansen2003reducing} representing state-of-the-art in evolutionary gradient estimation for continuous optimization.

\section{Flow Matching as Particle Swarm Optimization}

The Flow Matching (FM) framework and Particle Swarm Optimization (PSO) share a profound conceptual connection when viewed through the lens of population-based dynamics and vector-field-driven evolution. In FM, the initial noise distribution \( q(\mathbf{x}_0) \) (e.g., Gaussian) can be analogized to the initial particle swarm population in PSO, where each particle's position \( \mathbf{x}_i^0 \) is sampled from a predefined domain. The fitness of particles in PSO corresponds to the negative log-likelihood of the data distribution \( p(\mathbf{x}) \) in FM, guiding the optimization toward high-probability regions.

The velocity update rule in PSO, defined as:
\[
\mathbf{v}_i^{t+1} = w \mathbf{v}_i^t + c_1 r_1 (\mathbf{pbest}_i - \mathbf{x}_i^t) + c_2 r_2 (\mathbf{gbest} - \mathbf{x}_i^t),
\]
mirrors the role of the learned vector field \( \mathbf{v}_t(\mathbf{x}) \) in FM, which deterministically transports particles (or samples) from \( q(\mathbf{x}_0) \) to \( p(\mathbf{x}_1) \). Here, \( \mathbf{pbest}_i \) and \( \mathbf{gbest} \) act as local and global attractors, analogous to the conditional and marginal flow directions in FM that minimize the KL divergence between the model and target distributions. The stochasticity introduced by \( r_1, r_2 \) in PSO parallels the implicit randomness in FM's training objectives when approximating the vector field from finite data.

The convergence of PSO to an optimal solution distribution \( \mathbf{x}_i^T \sim p^*(\mathbf{x}) \) aligns with FM's asymptotic goal: the final particle distribution \( p(\mathbf{x}_1) \) matches the target data distribution, with higher fitness (likelihood) regions densely populated. This is achieved in FM through the ODE:
\[
\frac{d\mathbf{x}_t}{dt} = \mathbf{v}_t(\mathbf{x}_t), \quad \mathbf{x}_0 \sim q(\mathbf{x}_0),
\]
where the continuous-time dynamics generalize PSO's discrete updates, replacing heuristic velocity terms with a learned gradient flow. Notably, FM's deterministic trajectory construction avoids PSO's risk of oscillatory convergence, while PSO's population-based exploration offers a stochastic counterpart to FM's density-driven interpolation.

The synergy between these frameworks suggests potential cross-pollination: FM could benefit from PSO's multi-particle exploration for multimodal optimization, while PSO might adopt FM's continuous flow formalism to refine its convergence guarantees. This perspective positions FM as a gradient-aware, continuous-limit generalization of PSO, where the "swarm" evolves under a globally coherent vector field rather than local heuristic rules.

\section{Conclusion and Future Perspective}

This paper has briefly revisited two distinct fields—generative models and evolutionary computation—and initiated an exploration of the potential connections between flow matching and particle swarm optimization. The preliminary analogies drawn here open several promising directions for future research. First, a comprehensive investigation of the mathematical duality between flow matching models and swarm intelligence algorithms could establish direct theoretical links, potentially revealing deeper symmetries in their optimization mechanisms. Second, interpreting flow matching models through the lens of swarm intelligence may inspire novel algorithms, such as designing swarm-based methods where particle dynamics are governed by learned vector fields instead of heuristic rules. Third, a particularly promising direction lies in employing flow matching models to solve optimization problems, including both single-objective and multi-objective formulations, where the continuous-time dynamics could provide more efficient convergence properties compared to traditional evolutionary approaches. Fourth, revisiting the dynamical processes of swarm intelligence from a flow matching perspective could allow these methods to be approximated as Markov processes guided by ordinary differential equations (ODEs), potentially leading to more rigorous convergence analyses. Finally, a systematic review of the mathematical structures underlying existing swarm and evolutionary algorithms, with fundamental classification based on their ODE representations, could unify these approaches within a common theoretical framework. Such advancements would not only bridge the gap between these two fields but also potentially yield new hybrid methods combining the strengths of both paradigms—the gradient-aware continuous flows of modern generative models with the population-based exploration of bio-inspired optimization, particularly for complex optimization landscapes where current methods show limitations.

\bibliographystyle{unsrt}
\bibliography{ref}
\end{document}